\documentclass[10pt,twocolumn,letterpaper]{article}

\usepackage{wacv}

\usepackage{makecell}
\usepackage{multirow}
\usepackage{multicol}
\usepackage{booktabs}
\usepackage{tabularx}
\usepackage{pifont}
\usepackage{xcolor}
\usepackage{colortbl}
\newcolumntype{Y}{>{\centering\arraybackslash}X}
\newcolumntype{L}{>{\raggedright\arraybackslash}X}
\definecolor{oursbg}{RGB}{235,237,246}
\newcommand{\ours}{\rowcolor{oursbg}}
\definecolor{ForestGreen}{cmyk}{0.91,0,0.88,0.12}
\definecolor{BrickRed}{cmyk}{0,0.89,0.94,0.28}
\newcommand{\cmark}{{\color{ForestGreen!80}\ding{51}}}
\newcommand{\xmark}{{\color{BrickRed!80}\ding{55}}}
\newcommand{\pmbox}[1]{\raisebox{-.35ex}{\fontsize{5.5pt}{6.5pt}\selectfont$\pm$#1}}
\newcommand{\val}[2]{#1\pmbox{#2}}
\newcommand{\vb}[2]{\textbf{#1}\pmbox{#2}}
\newcommand{\vu}[2]{\underline{#1}\pmbox{#2}}
\newcommand{\gs}[1]{\textcolor{black!45}{#1}}

\definecolor{wacvblue}{rgb}{0.21,0.49,0.74}
\usepackage[pagebackref,breaklinks,colorlinks,allcolors=wacvblue]{hyperref}

\def\wacvPaperID{1427}
\def\confName{WACV}
\def\confYear{2027}

\title{
Complexity-Aware Surgical Video Reasoning Segmentation with Digital Twin and Context Memory
}

\author{Yiqing Shen \qquad Mathias Unberath\\
Department of Computer Science, Johns Hopkins University\\
{\tt\small \{yshen92, unberath\}@jhu.edu}
}

\begin{document}
\maketitle
\begin{abstract}

Conventional surgical segmentation relies on closed-set categories or explicit text queries, which cannot resolve implicit queries that require reasoning about workflows, instrument functions, and temporal relationships.
To address these limitations, we propose SurgReasonSeg, a complexity-aware method that brings reasoning segmentation over digital twin (DT) representations to surgical video.
Specifically, SurgReasonSeg constructs DT representations from surgical foundation models and trains two low-rank adapters (LoRA) of one large language model with reinforcement learning (RL), each specialized for a different query complexity.
A lightweight router, also trained with RL, selects between the two adapters to balance latency against accuracy.
SurgReasonSeg also sets the temporal context with an adaptive sliding window conditioned on query complexity and workflow phase, and retrieves relevant historical cases from a surgical context memory by workflow similarity and procedural relevance.
We further introduce EndoVis-ReasonSeg, a surgical reasoning segmentation benchmark with 1,460 training and 961 test samples across semantic, spatial, and temporal categories.
On EndoVis-ReasonSeg, SurgReasonSeg improves $\mathcal{J}\&\mathcal{F}$ over the strongest baseline, ReSurgSAM2, by 0.178, 0.237, and 0.297 on semantic, spatial, and temporal queries, and it also achieves the best results on the Ref-EndoVis17 and Ref-EndoVis18 referring segmentation benchmarks.
A backbone-matched attribution protocol separates the contribution of the upstream segmenter from that of the reasoning modules.
Our method runs at 18.8 FPS averaged over the query distribution, enabling real-time reasoning segmentation in surgical videos.

\end{abstract}

\section{Introduction}

Surgical autonomy requires precise segmentation of anatomical structures and instruments in endoscopic videos~\cite{attanasio2021autonomy}, yet surgeons often express their intent through functional and contextual descriptions tied to the ongoing procedure rather than through explicit anatomical labels~\cite{shen2025operating}.
Conventional segmentation methods rely on pre-defined closed-set categories~\cite{raso}, restricting them to a fixed vocabulary that cannot accommodate such intent-driven communication.
Referring segmentation~\cite{referringsegmentation} partially mitigates this by identifying objects through explicit text queries~\cite{zhou2023text,resurgsam2,rsvis}, yet still requires the surgeon to name the target.
In practice, many surgical tasks require interpreting implicit queries that involve reasoning about workflows, instrument functions, and temporal procedural relationships~\cite{surgraw}.
For instance, rather than ``\textit{the grasper}'', a surgeon may ask for ``\textit{the instrument retracting tissue for dissection}'', expecting the target to be inferred from workflow context and instrument roles, a task that reasoning segmentation addresses through multi-step inference~\cite{visa}.

Existing reasoning segmentation methods such as LISA~\cite{lisa} and VISA~\cite{visa} pair multimodal large language models (LLMs) with segmentation decoders, yet are trained on general-domain data and lack surgical knowledge~\cite{shen2025operating}.
While fine-tuning on surgical reasoning segmentation data could offer a partial solution, it requires costly annotations and yields a model whose knowledge is fixed at training time.
Existing methods also cannot retrieve relevant historical context from long procedures, either processing all information uniformly or discarding past observations~\cite{ding2025mosformer,jin2021temporal}.
Finally, surgical queries vary in reasoning complexity, from identifying an object to understanding the workflow~\cite{xue2025omni}, yet existing methods allocate the same computation to every query and spend unnecessary latency on simple ones.

To address these limitations, we propose SurgReasonSeg, an online video reasoning segmentation method built on digital twin (DT) representations~\cite{jit,position,dtr1,fastreasonseg}, which serve as intermediate abstractions between raw video and reasoning.
SurgReasonSeg constructs the DT with surgical foundation models such as SurgSAM2~\cite{surgicalsam2} and RASO~\cite{raso}, incorporating surgical domain knowledge directly into the representation.
The DT-as-JSON encoding and the reduction of segmentation to index retrieval are inherited from prior DT reasoning segmentation methods~\cite{jit,shen2025operating,dtr1,fastreasonseg}, not proposed here.
Tab.~\ref{tab:positioning} lists what is inherited and what is new, and Tab.~\ref{tab:attribution} quantifies the contribution of each part.
To handle varying query complexity, we train two low-rank adapters (LoRA)~\cite{lora} on the same base LLM with reinforcement learning (RL)~\cite{grpo}, one producing extended reasoning chains and the other concise ones.
A lightweight router, also trained with RL, selects between the two adapters to balance latency against accuracy.
We also propose an adaptive sliding window that adjusts the length of the temporal context to the query complexity and the workflow phase, so that the LLM receives only as much context as the query requires.
To reuse experience accumulated across cases, we introduce a surgical context memory that stores past DT representations together with their queries and outcomes.
At inference, the memory retrieves past cases by workflow similarity and procedural relevance and supplies them to the LLM as in-context examples.

The major contributions are threefold, summarized as follows.
First, we propose SurgReasonSeg, an online video reasoning segmentation method that adapts DT-based reasoning segmentation to surgical video by constructing the DT from surgical foundation models.
Specifically, two RL-trained LoRA adapters handle queries of different complexity, and a router selects between them at inference.
Second, we introduce an adaptive sliding window that sets the temporal context from the query complexity and the surgical workflow phase, and a surgical context memory that retrieves past cases by workflow similarity and procedural relevance.
Finally, we contribute EndoVis-ReasonSeg, a surgical reasoning segmentation benchmark for training and evaluation.
We also report a backbone-matched attribution study that separates the contribution of the upstream segmenter from that of the reasoning modules, together with a template-disjoint evaluation that tests for template-level leakage.

\section{Related Work}

\noindent\textbf{Reasoning Segmentation.}
Reasoning segmentation aims to identify and segment objects from implicit text queries that do not name the target explicitly.
LISA~\cite{lisa} attaches a segmentation decoder to a multimodal LLM through a learned \texttt{[SEG]} token, GSVA~\cite{gsva} generalizes it to multiple and empty targets, CoReS~\cite{cores} interleaves chain-of-thought reasoning with mask decoding, and SegZero~\cite{segzero} trains the reasoning chain through RL.
For videos, VISA~\cite{visa} selects a keyframe before propagating masks, while video LLMs~\cite{videollama,videochatgpt,llamavid} compress frames into token sequences.
However, these methods encode the video as visual tokens and are trained on general-domain corpora, leaving the surgical vocabulary and procedural priors that implicit surgical queries depend on unmodeled.

\noindent\textbf{Digital Twin Representations.}
Digital twins provide a structured, symbolic abstraction of a physical scene~\cite{dtreview,ditto} that recent work uses as the interface between perception and LLM reasoning~\cite{position}.
In surgery, DT scene representations built from foundation models support task-level automation~\cite{dtauto} and robust phase recognition~\cite{dtphase}, and Shen \etal~\cite{shen2025operating} first perform reasoning segmentation over DTs for operating room workflow analysis.
The DT-as-JSON encoding with masks stored as file paths, together with the reduction of segmentation to index retrieval, was introduced by JiT~\cite{jit} and subsequently refined for general visual reasoning~\cite{dtr1} and for efficiency~\cite{fastreasonseg}.
Outside vision, generative transformers have also modeled structured sequences such as entity trajectories~\cite{trackgpt}.
SurgReasonSeg inherits this DT formulation, and Tab.~\ref{tab:positioning} lists which of its components are inherited, adapted, or new.

\noindent\textbf{Surgical Video Segmentation.}
Surgical segmentation has progressed from closed-set categories~\cite{raso,surgicalsam2} to referring formulations, where RSVIS~\cite{rsvis} exploits video-level consistency and ReSurgSAM2~\cite{resurgsam2} couples a surgical encoder with SAM2 tracking to reach real-time speed.
Yet both require the surgeon to name the target, and neither reasons over workflow phases or instrument function.
Temporal context has been modeled through memory banks for phase recognition~\cite{ding2025mosformer,jin2021temporal} and adaptive retrieval for long videos~\cite{xue2025omni}, yet neither is conditioned on query complexity.

\begin{table}[t!]
\centering
\caption{\textbf{SurgReasonSeg inherits the DT encoding and adds four components.}
Adapted: a prior component modified for surgery.}
\label{tab:positioning}
\footnotesize
\setlength{\tabcolsep}{3pt}
\begin{tabularx}{\columnwidth}{@{}Lll@{}}
\toprule
Component & Origin & Status \\
\midrule
DT-as-JSON and index retrieval & \cite{jit,dtr1,fastreasonseg} & Inherited \\
Reasoning segmentation on surgical DTs & \cite{shen2025operating} & Inherited \\
Surgical DT from foundation models & \cite{dtauto,dtphase} & Adapted \\
\midrule
Two LoRA adapters with an RL router & This paper & New \\
Adaptive sliding window & This paper & New \\
Surgical context memory & This paper & New \\
EndoVis-ReasonSeg benchmark & This paper & New \\
\bottomrule
\end{tabularx}
\end{table}

\section{Methods}
\label{sec:method}

\begin{figure*}[t!]
\centering
\includegraphics[width=\linewidth]{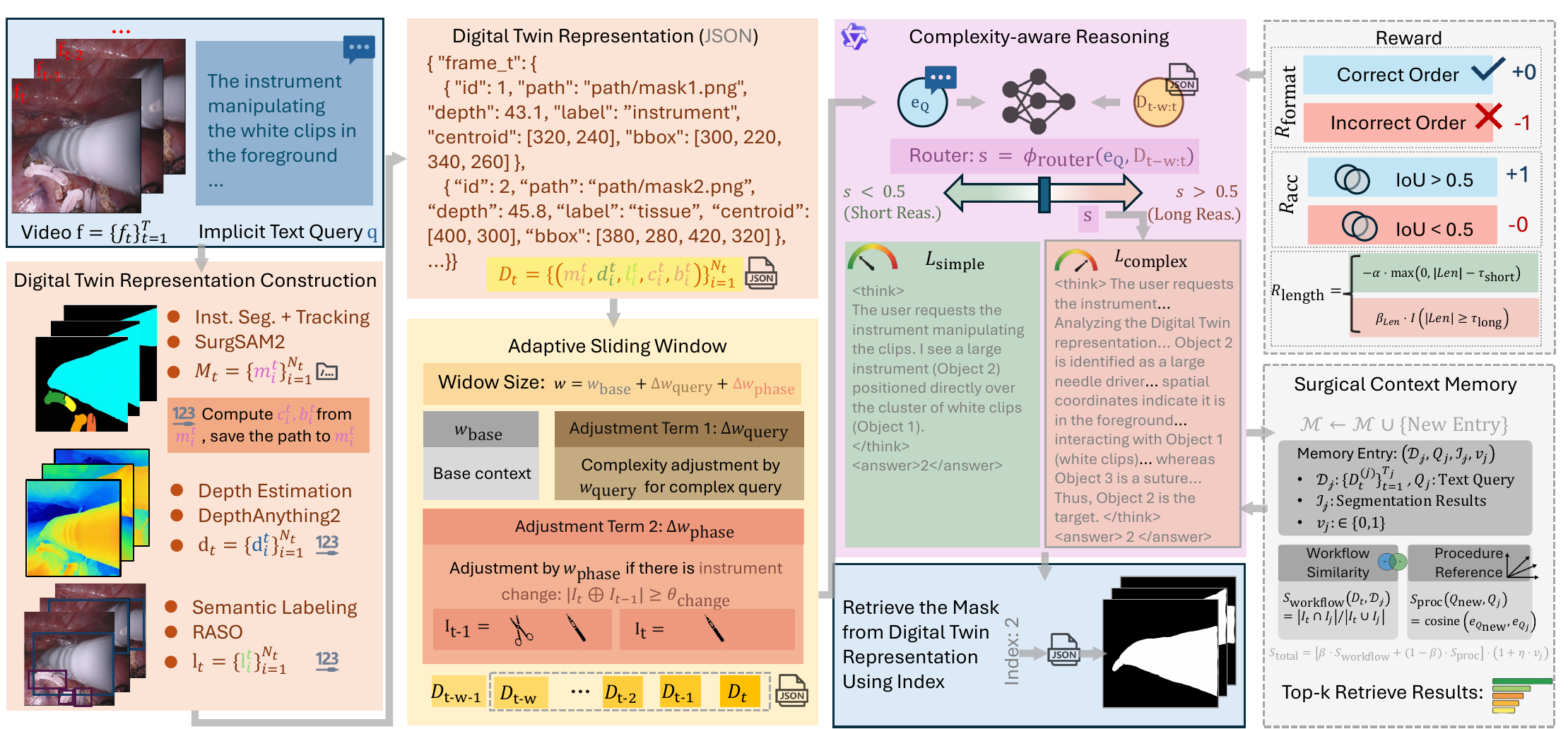}
\caption{
Overview of SurgReasonSeg.
Given a surgical video and an implicit text query, our method constructs DT representations from surgical foundation models, applies an adaptive sliding window for temporal context selection, and routes each query to the LoRA adapter matched to its complexity, which generates the reasoning chain.
}
\label{fig:method}
\end{figure*}

As shown in Fig.~\ref{fig:method}, SurgReasonSeg combines four components, namely DT construction, an adaptive sliding window, a surgical context memory, and complexity-aware reasoning with a router.
The window selects the temporal context, the memory supplies similar past cases as in-context examples, and the router sends each query to one of two LoRA adapters, which generates a reasoning chain followed by the index of the target mask.

\noindent\textbf{Digital Twin Construction.}
Given a surgical video clip with frames $\{f_t\}_{t=1}^T$ and text query $Q$, SurgReasonSeg first constructs the DT representation $D_t$ at each time step $t$ to capture the surgical scene.
First, we apply SurgSAM2~\cite{surgicalsam2} for instrument and tissue instance segmentation, generating instance masks $M_t = \{m_i^t\}_{i=1}^{N_t}$ where each $m_i^t$ is the binary mask of object $i$ in frame $t$.
The tracking of SurgSAM2 keeps instance identities consistent across frames.
Second, we estimate depth with DepthAnything2~\cite{depthanything} and compute the average depth $d_i^t$ of each instance.
Third, RASO~\cite{raso} assigns each instance a surgical semantic label $l_i^t$.
The DT representation at frame $t$ is formalized as $D_t = \{(m_i^t, d_i^t, l_i^t, c_i^t, b_i^t)\}_{i=1}^{N_t}$, where each tuple contains the mask, average depth, semantic label, centroid $c_i^t$, and bounding box coordinates $b_i^t$ for object $i$.
The centroid and bounding box, both derived from the mask, allow the LLM to reason about spatial relations without reading the pixel-level masks.
Following prior work~\cite{jit,shen2025operating,dtr1,fastreasonseg}, we serialize $D = \{D_t\}$ as JSON that stores each mask as a file path instead of a pixel array, which turns reasoning segmentation into mask retrieval in which the LLM identifies the target object by its JSON index.

\noindent\textbf{Complexity-Aware Reasoning with Router.}
\label{sec:router}
Following prior work~\cite{dtr1,fastreasonseg}, our LLM takes the DT representation $D_{t-w:t}$ from the sliding window and the implicit text query $Q$ as input to generate a two-component rollout.
The rollout first gives a reasoning chain within \texttt{<think>} and \texttt{</think>} tokens that analyzes object attributes and spatial relations in the DT representation.
It then gives the JSON index of the target object within \texttt{<answer>} and \texttt{</answer>} tokens.
We train the LoRA adapters on the LLM through RL with the reward $R = R_{\text{format}} + R_{\text{acc}}$.
The format reward $R_{\text{format}}$ assigns $0$ to a well-formed rollout and $-1$ otherwise, so that malformed outputs are penalized without format compliance being rewarded on its own.
The accuracy reward $R_{\text{acc}}$ equals $1$ when the IoU between the retrieved mask and the ground truth exceeds $0.5$ and $0$ otherwise.
The IoU criterion is applied per referent, and each query is authored so that exactly one DT instance satisfies the description at the queried frame.
Set-like phrasings are rewritten at the template level to designate a single instance, for example ``\textit{the primary grasping tool}'' rather than ``\textit{instruments active during dissection phase}''.
To handle varying query complexity, we train two LoRA adapters~\cite{lora} on the same base LLM that differ in the length of their reasoning chain within the \texttt{<think>} and \texttt{</think>} tokens.
The adapter $L_{\text{complex}}$ generates extended reasoning for complex queries, while $L_{\text{simple}}$ produces concise reasoning for straightforward ones.
Each adapter adds an efficiency reward $R_{\text{length}}$ to $R$, defined as
\begin{equation}
\label{eq:length}
    R_{\text{length}} = \begin{cases}
-\alpha \cdot \max(0, |Len| - \tau_{\text{short}}) & \text{for } L_{\text{simple}} \\
\beta_{Len} \cdot \mathbb{I} (|Len| \geq \tau_{\text{long}}) & \text{for } L_{\text{complex}}
\end{cases}
\end{equation}
where $|Len|$ denotes the length of the reasoning chain, $\tau_{\text{short}}$ and $\tau_{\text{long}}$ are the length thresholds, $\alpha$ and $\beta_{Len}$ are the weighting coefficients, and $\mathbb{I} (|Len| \geq \tau_{\text{long}})$ is an indicator function.
To select between the two adapters at inference, we then train a lightweight router network $\phi_{\text{router}}$, implemented as Qwen3-0.6B~\cite{qwen3}, with RL while both adapters stay frozen.
The router takes the query embedding $e_Q$ and the DT window $D_{t-w:t}$ as input and produces a selection score $s = \phi_{\text{router}}(e_Q, D_{t-w:t})$, where $s > 0.5$ activates $L_{\text{complex}}$ and $s \leq 0.5$ activates $L_{\text{simple}}$.
The router is trained with the reward function $R_{\text{router}} = R_{\text{acc}} - \lambda \cdot T_{\text{latency}}$,
where $T_{\text{latency}}$ measures the inference time of the selected adapter and $\lambda$ balances the segmentation accuracy against computational efficiency.
This reward encourages the router to select $L_{\text{simple}}$ for straightforward queries to minimize latency while choosing $L_{\text{complex}}$ only when extended reasoning is necessary for accuracy.
No query is labeled simple or complex by hand, and the specialization is induced only by $R_{\text{length}}$ with $\tau_{\text{short}}=50$ and $\tau_{\text{long}}=200$ tokens, which are reasoning-chain length thresholds rather than complexity labels, and the router learns its decision boundary from $R_{\text{router}}$ alone.
We define the ``no Router'' configuration used in the ablation as a single LoRA adapter trained on all queries with $R = R_{\text{format}} + R_{\text{acc}}$ and without the length reward $R_{\text{length}}$, so that removing the router also removes the complexity specialization it selects between.

\begin{figure}[t!]
\centering
\includegraphics[width=\linewidth]{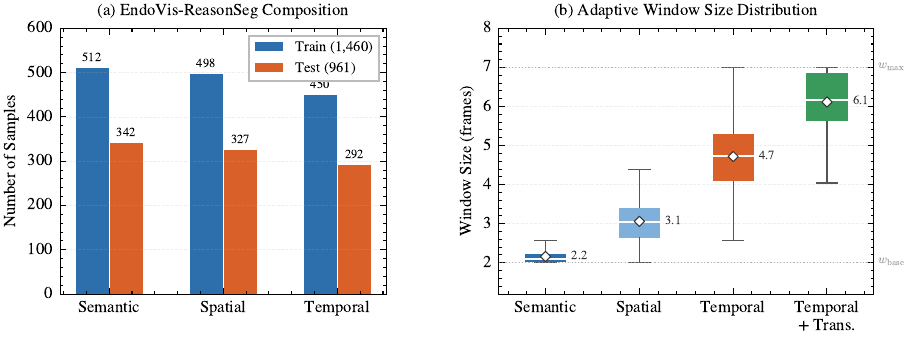}
\caption{
Dataset composition and adaptive window sizes.
The left panel shows the number of EndoVis-ReasonSeg samples per reasoning category in the training and test splits, and the right panel shows the distribution of adaptive sliding window sizes across query types, with diamond markers indicating means.
Temporal queries during phase transitions receive the largest windows (mean 6.1 frames), while semantic queries remain close to $w_{\text{base}}=2$.
}
\label{fig:dataset}
\end{figure}

\noindent\textbf{Adaptive Sliding Window.}
The adaptive sliding window adjusts the temporal context length of DT representations fed to the LLM based on query complexity and surgical workflow phases.
At each time step $t$, the window size $w$ that determines $D_{t-w:t}$ is computed as $w = w_{\text{base}} + \Delta w_{\text{query}} + \Delta w_{\text{phase}}$,
where $w_{\text{base}}$ provides a minimal temporal context.
The query complexity adjustment $\Delta w_{\text{query}}$ extends the window when the router identifies a complex query, namely $\Delta w_{\text{query}} = w_{\text{query}} \cdot \mathbb{I}(s > 0.5)$.
Because this adjustment depends on $s$, the router reads the window of $w_{\text{base}} + \Delta w_{\text{phase}}$ frames, both known before routing, and only the selected adapter reads the full window.
The workflow phase adjustment $\Delta w_{\text{phase}}$ expands the context during surgical transitions, which are detected from changes in the active instruments recorded in the DT.
Specifically, $\Delta w_{\text{phase}} = w_{\text{phase}}$ if $|I_t \oplus I_{t-1}| \ge \theta_{\text{change}}$ and $0$ otherwise, where $I_t$ denotes the set of active instruments in frame $t$ extracted from the DT representation.
The operator $\oplus$ is the symmetric difference, which returns the instruments present in exactly one of the two frames, and $\theta_{\text{change}}$, set to 1, is the minimum number of instrument changes that triggers a workflow phase transition.

\begin{table*}[t!]
\centering
\caption{\textbf{SurgReasonSeg achieves the best $\mathcal{J}$, $\mathcal{F}$, and $\mathcal{J}\&\mathcal{F}$ on all three reasoning categories of EndoVis-ReasonSeg.}
Reas./Ref.: reasoning/referring methods; $\pm$: standard deviation over three random seeds; the FPS of SurgReasonSeg includes DT construction and is averaged over routing.
Bold: best; underline: second best.}
\label{tab:endovis_reasonseg_results}
\footnotesize
\setlength{\tabcolsep}{2pt}
\begin{tabularx}{\textwidth}{@{}llYYYYYYYYYY@{}}
\toprule
& & \multicolumn{3}{c}{Semantic} & \multicolumn{3}{c}{Spatial} & \multicolumn{3}{c}{Temporal} & \\
\cmidrule(lr){3-5}\cmidrule(lr){6-8}\cmidrule(lr){9-11}
Method & Type & $\mathcal{J}\uparrow$ & $\mathcal{F}\uparrow$ & $\mathcal{J}\&\mathcal{F}\uparrow$ & $\mathcal{J}\uparrow$ & $\mathcal{F}\uparrow$ & $\mathcal{J}\&\mathcal{F}\uparrow$ & $\mathcal{J}\uparrow$ & $\mathcal{F}\uparrow$ & $\mathcal{J}\&\mathcal{F}\uparrow$ & FPS$\uparrow$ \\
\midrule
LISA~\cite{lisa} & Reas. & \val{.556}{.031} & \val{.568}{.024} & \val{.562}{.025} & \val{.201}{.042} & \val{.213}{.035} & \val{.207}{.036} & \val{.238}{.039} & \val{.251}{.033} & \val{.245}{.034} & 1.33 \\
VISA~\cite{visa} & Reas. & \val{.528}{.023} & \val{.547}{.021} & \val{.538}{.019} & \val{.457}{.032} & \val{.481}{.028} & \val{.469}{.027} & \val{.319}{.037} & \val{.338}{.031} & \val{.329}{.032} & 1.87 \\
LLM-Seg~\cite{llmseg} & Reas. & \val{.387}{.034} & \val{.394}{.031} & \val{.391}{.029} & \val{.245}{.043} & \val{.258}{.038} & \val{.252}{.037} & \val{.176}{.048} & \val{.189}{.041} & \val{.183}{.042} & 0.29 \\
SegZero~\cite{segzero} & Reas. & \val{.509}{.025} & \val{.526}{.022} & \val{.518}{.021} & \val{.398}{.033} & \val{.421}{.029} & \val{.410}{.028} & \val{.312}{.038} & \val{.331}{.032} & \val{.322}{.033} & 0.11 \\
CoReS~\cite{cores} & Reas. & \val{.560}{.022} & \val{.580}{.019} & \val{.570}{.018} & \val{.439}{.028} & \val{.463}{.024} & \val{.451}{.023} & \val{.351}{.033} & \val{.372}{.029} & \val{.362}{.028} & 0.27 \\
JiT~\cite{jit} & Reas. & \val{.548}{.024} & \val{.565}{.021} & \val{.557}{.020} & \val{.452}{.029} & \val{.476}{.025} & \val{.464}{.024} & \vu{.364}{.032} & \vu{.385}{.028} & \vu{.375}{.027} & 0.03 \\
RSVIS~\cite{rsvis} & Ref. & \val{.542}{.023} & \val{.559}{.020} & \val{.551}{.019} & \val{.391}{.031} & \val{.412}{.027} & \val{.402}{.026} & \val{.296}{.035} & \val{.315}{.030} & \val{.306}{.029} & \underline{22.1} \\
ReSurgSAM2~\cite{resurgsam2} & Ref. & \vu{.584}{.019} & \vu{.601}{.017} & \vu{.593}{.016} & \vu{.468}{.024} & \vu{.492}{.021} & \vu{.480}{.020} & \val{.347}{.028} & \val{.367}{.025} & \val{.357}{.024} & \textbf{61.2} \\
\midrule
\ours SurgReasonSeg (Ours) & Reas. & \vb{.762}{.013} & \vb{.779}{.011} & \vb{.771}{.010} & \vb{.704}{.018} & \vb{.729}{.014} & \vb{.717}{.013} & \vb{.641}{.021} & \vb{.667}{.018} & \vb{.654}{.017} & 18.8 \\
\bottomrule
\end{tabularx}
\end{table*}

\noindent\textbf{Surgical Context Memory.}
We maintain a surgical context memory bank $\mathcal{M}$ that stores DT representations alongside queries and segmentation results.
The bank retains only DT attributes, mask paths, and queries instead of raw frames, which limits pixel-level exposure, though these attributes are still derived from patient recordings and require the same governance as the source videos.
Each memory entry is a tuple $(\mathcal{D}_j, Q_j, \mathcal{I}_j, v_j)$ where $\mathcal{D}_j = \{D_t^{(j)}\}_{t=1}^{T_j}$ represents the complete DT sequence for video $j$, $Q_j$ is the text query, $\mathcal{I}_j$ contains segmentation results, and $v_j \in \{0, 1\}$ indicates unsuccessful or successful segmentation.
The label $v_j$ is produced by an external VLM judge, instantiated as GPT-4o, that is shown the retrieved mask overlaid on the frame together with the query and returns a binary verdict on whether the mask satisfies the description.
Tab.~\ref{tab:design} of the supplementary replaces $v_j$ by the ground-truth IoU to measure how much memory quality depends on the judge.
The memory stores both successful and unsuccessful results, updated as $\mathcal{M} \leftarrow \mathcal{M} \cup \{(\mathcal{D}_j, Q_j, \mathcal{I}_j, v_j)\}$.
To avoid leakage, the bank is built only from training videos and frozen during evaluation, so the update rule applies only at deployment and no test case can retrieve another.
For a query $Q_{\text{new}}$ at frame $t$, we retrieve relevant historical context by the workflow similarity $S_{\text{workflow}}(D_t, \mathcal{D}_j) = |I_t \cap I_j| / |I_t \cup I_j|$, where $I_t$ and $I_j$ are sets of instruments.
We also compute the procedural relevance $S_{\text{proc}}(Q_{\text{new}}, Q_j) = \text{cosine}(e_{Q_{\text{new}}}, e_{Q_j})$, the cosine similarity between the query embeddings.
The retrieval score incorporates quality judgment: $S_{\text{total}} = [\beta \cdot S_{\text{workflow}} + (1-\beta) \cdot S_{\text{proc}}] \cdot (1 + \eta \cdot v_j)$, where $\eta$ weights successful examples while retaining unsuccessful ones as contrasting references.
We retrieve the top-$k$ memories and prepend their temporal segments with corresponding results as few-shot examples, namely $D'_{t-w:t} = [(\mathcal{D}_{j_1}^{\text{seg}}, \mathcal{I}_{j_1}, v_{j_1}); \ldots; (\mathcal{D}_{j_k}^{\text{seg}}, \mathcal{I}_{j_k}, v_{j_k}); D_{t-w:t}]$, where each tuple provides the DT representation, segmentation result, and quality label, so the LLM sees both successful and unsuccessful cases as in-context examples.

\noindent\textbf{Dataset Construction.}
We construct a surgical reasoning segmentation dataset called EndoVis-ReasonSeg using the videos and instance segmentation masks from Ref-EndoVis17, Ref-EndoVis18~\cite{resurgsam2}, and RSVIS~\cite{rsvis}.
We segment each video sequence into overlapping clips of 20 to 40 consecutive frames, where clip boundaries align with surgical workflow transitions detected through changes in active instruments.
For each clip, we manually create implicit text queries spanning three reasoning categories: (1) semantic queries requiring understanding of instrument roles without explicit names (\eg, ``\textit{the tool used for tissue retraction}''), (2) spatial queries demanding relative positioning and anatomical context (\eg, ``\textit{the instrument closest to the liver}''), and (3) temporal queries involving procedural phases and workflow understanding (\eg, ``\textit{the instrument active during the dissection phase}'').
Each sample consists of a video clip, an implicit query, and the corresponding ground-truth mask.
The training set contains 7 video sequences from EndoVis17 (215 clips, 873 samples) and 11 video sequences from EndoVis18 (164 clips, 587 samples), totaling 18 video sequences.
The test set includes 3 video sequences from EndoVis17 (105 clips, 365 samples) and 4 video sequences from EndoVis18 (70 clips, 596 samples).
At 2,421 query-mask pairs, the benchmark is about twice the size of ReasonSeg~\cite{lisa}, the reference benchmark for this task, which contains 1,218 image-query pairs.
Training and test videos are disjoint at the sequence level.
Since the 47 query templates are shared across splits, we also build a template-disjoint split with 10 templates held out from training, reported in Tab.~\ref{tab:attribution}, to test for template-level memorization.

Fig.~\ref{fig:dataset}(a) visualizes the dataset composition across the three reasoning categories.

\section{Experiments}
\label{sec:exp}

\begin{figure}[t!]
\centering
\includegraphics[width=\linewidth]{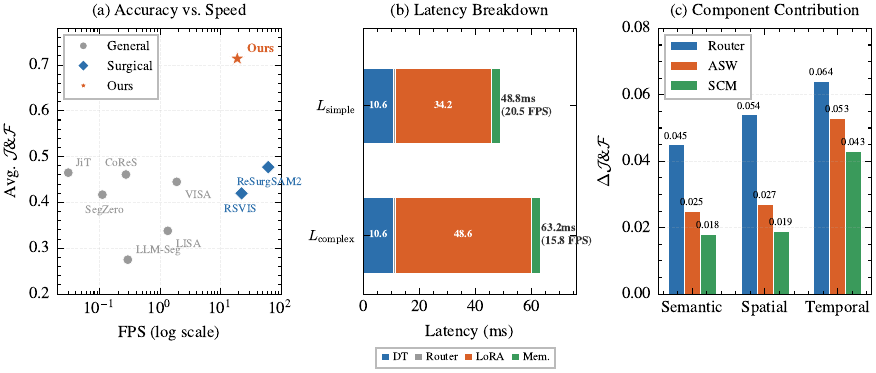}
\caption{
Accuracy, latency, and component analysis.
The left panel plots the average $\mathcal{J}\&\mathcal{F}$ over the three reasoning categories against FPS for all methods, where SurgReasonSeg (star) gives the best balance between reasoning accuracy and speed.
The middle panel breaks down the latency of the $L_{\text{simple}}$ and $L_{\text{complex}}$ inference paths per component on a single RTX 4090 GPU.
The right panel shows the drop in $\mathcal{J}\&\mathcal{F}$ when the router, ASW, or SCM is removed, and each of the three components has its largest effect on temporal reasoning.
}
\label{fig:analysis}
\end{figure}

\begin{table*}[t!]
\centering
\caption{\textbf{SurgReasonSeg achieves the best $\mathcal{J}\&\mathcal{F}$ on all three referring segmentation settings.}
$\pm$: standard deviation over three random seeds; bold: best; underline: second best.}
\label{tab:referring_segmentation_results}
\footnotesize
\setlength{\tabcolsep}{2pt}
\begin{tabularx}{\textwidth}{@{}lYYYYYYYYY@{}}
\toprule
& \multicolumn{3}{c}{Ref-EndoVis17~\cite{resurgsam2} (tool)} & \multicolumn{3}{c}{Ref-EndoVis18~\cite{resurgsam2} (tool)} & \multicolumn{3}{c}{Ref-EndoVis18~\cite{resurgsam2} (tissue)} \\
\cmidrule(lr){2-4}\cmidrule(lr){5-7}\cmidrule(lr){8-10}
Method & $\mathcal{J}\&\mathcal{F}\uparrow$ & $\mathcal{J}\uparrow$ & $\mathcal{F}\uparrow$ & $\mathcal{J}\&\mathcal{F}\uparrow$ & $\mathcal{J}\uparrow$ & $\mathcal{F}\uparrow$ & $\mathcal{J}\&\mathcal{F}\uparrow$ & $\mathcal{J}\uparrow$ & $\mathcal{F}\uparrow$ \\
\midrule
LISA~\cite{lisa} & \val{.388}{.032} & \val{.420}{.029} & \val{.355}{.034} & \val{.293}{.041} & \val{.284}{.043} & \val{.303}{.037} & \val{.237}{.044} & \val{.244}{.042} & \val{.229}{.045} \\
VISA~\cite{visa} & \val{.507}{.022} & \val{.479}{.026} & \val{.535}{.019} & \val{.431}{.028} & \val{.398}{.031} & \val{.464}{.024} & \val{.384}{.031} & \val{.354}{.033} & \val{.414}{.027} \\
LLM-Seg~\cite{llmseg} & \val{.307}{.039} & \val{.177}{.047} & \val{.437}{.031} & \val{.177}{.048} & \val{.105}{.052} & \val{.249}{.041} & \val{.153}{.049} & \val{.103}{.053} & \val{.203}{.043} \\
SegZero~\cite{segzero} & \val{.468}{.027} & \val{.388}{.032} & \val{.548}{.022} & \val{.388}{.031} & \val{.349}{.034} & \val{.427}{.028} & \val{.349}{.033} & \val{.321}{.035} & \val{.377}{.030} \\
CoReS~\cite{cores} & \val{.495}{.023} & \val{.415}{.028} & \val{.575}{.019} & \val{.415}{.027} & \val{.377}{.030} & \val{.453}{.024} & \val{.377}{.029} & \val{.351}{.031} & \val{.403}{.026} \\
JiT~\cite{jit} & \val{.509}{.021} & \val{.441}{.026} & \val{.577}{.018} & \val{.429}{.027} & \val{.403}{.029} & \val{.455}{.023} & \val{.391}{.028} & \val{.361}{.031} & \val{.421}{.025} \\
RSVIS~\cite{rsvis} & \val{.612}{.018} & \val{.610}{.019} & \val{.615}{.017} & \val{.684}{.016} & \val{.683}{.017} & \val{.685}{.016} & \val{.657}{.019} & \val{.729}{.015} & \val{.585}{.022} \\
ReSurgSAM2~\cite{resurgsam2} & \vu{.777}{.013} & \vu{.777}{.014} & \vu{.777}{.013} & \vu{.806}{.011} & \vu{.809}{.012} & \vu{.803}{.011} & \vu{.751}{.016} & \vu{.809}{.012} & \vu{.693}{.019} \\
\midrule
\ours SurgReasonSeg (Ours) & \vb{.801}{.011} & \vb{.798}{.012} & \vb{.805}{.010} & \vb{.829}{.009} & \vb{.832}{.010} & \vb{.826}{.009} & \vb{.778}{.013} & \vb{.835}{.010} & \vb{.721}{.014} \\
\bottomrule
\end{tabularx}
\end{table*}

\begin{table*}[htbp!]
\centering
\caption{\textbf{Removing any component lowers $\mathcal{J}\&\mathcal{F}$, most on temporal queries.}
Router: routing between two LoRA adapters; ASW: adaptive sliding window; SCM: surgical context memory; $\pm$: standard deviation over three random seeds.
Bold: best; underline: second best; FPS is a cost and is not ranked.}
\label{tab:ablation_study}
\footnotesize
\setlength{\tabcolsep}{2pt}
\begin{tabularx}{\textwidth}{@{}ccccYYYYYYYYYc@{}}
\toprule
& & & & \multicolumn{3}{c}{Semantic} & \multicolumn{3}{c}{Spatial} & \multicolumn{3}{c}{Temporal} & \\
\cmidrule(lr){5-7}\cmidrule(lr){8-10}\cmidrule(lr){11-13}
DT & Router & ASW & SCM & $\mathcal{J}\uparrow$ & $\mathcal{F}\uparrow$ & $\mathcal{J}\&\mathcal{F}\uparrow$ & $\mathcal{J}\uparrow$ & $\mathcal{F}\uparrow$ & $\mathcal{J}\&\mathcal{F}\uparrow$ & $\mathcal{J}\uparrow$ & $\mathcal{F}\uparrow$ & $\mathcal{J}\&\mathcal{F}\uparrow$ & FPS \\
\midrule
\xmark & \xmark & \xmark & \xmark & \val{.587}{.031} & \val{.601}{.028} & \val{.594}{.027} & \val{.445}{.038} & \val{.468}{.033} & \val{.457}{.033} & \val{.382}{.041} & \val{.405}{.037} & \val{.394}{.036} & 1.5 \\
\cmark & \xmark & \cmark & \cmark & \val{.718}{.021} & \val{.734}{.018} & \val{.726}{.017} & \val{.651}{.024} & \val{.675}{.021} & \val{.663}{.020} & \val{.578}{.027} & \val{.602}{.023} & \val{.590}{.022} & 12.5 \\
\cmark & \cmark & \xmark & \cmark & \val{.738}{.018} & \val{.754}{.015} & \val{.746}{.014} & \val{.678}{.022} & \val{.702}{.019} & \val{.690}{.018} & \val{.589}{.025} & \val{.613}{.022} & \val{.601}{.021} & 17.8 \\
\cmark & \cmark & \cmark & \xmark & \vu{.745}{.015} & \vu{.761}{.013} & \vu{.753}{.012} & \vu{.685}{.020} & \vu{.710}{.017} & \vu{.698}{.016} & \vu{.598}{.023} & \vu{.623}{.020} & \vu{.611}{.019} & 20.0 \\
\midrule
\ours \cmark & \cmark & \cmark & \cmark & \vb{.762}{.013} & \vb{.779}{.011} & \vb{.771}{.010} & \vb{.704}{.018} & \vb{.729}{.014} & \vb{.717}{.013} & \vb{.641}{.021} & \vb{.667}{.018} & \vb{.654}{.017} & 18.8 \\
\bottomrule
\end{tabularx}
\end{table*}

\noindent\textbf{Implementation Details.}
We implement SurgReasonSeg with Qwen3-4B~\cite{qwen3} as the base LLM, training on 8 NVIDIA RTX 4090 GPUs using LoRA~\cite{lora} with rank 64 and scaling factor 128.
Both LoRA adapters and the router are trained with GRPO~\cite{grpo} in TRL at a learning rate of $4 \times 10^{-5}$, and the router reward weights latency with $\lambda=0.3$.
The two adapters are trained sequentially, $L_{\text{complex}}$ then $L_{\text{simple}}$ for 3 epochs each, followed by the router for 2 epochs with frozen LoRA weights, and reward variance drops below 0.05 after the first epoch.

For Eq.~\eqref{eq:length} we set $\alpha=0.01$, $\beta_{Len}=0.3$, $\tau_{\text{short}}=50$, $\tau_{\text{long}}=200$ tokens; the window uses $w_{\text{base}}=2$, $w_{\text{query}}=2$, $w_{\text{phase}}=3$, $\theta_{\text{change}}=1$, and the memory $\beta=0.6$, $\eta=0.5$, $k=3$.
The bank holds 500 entries, evicting the lowest-scoring one when full.
Retrieval uses FAISS and takes 1.2ms for 500 entries and 2.1ms for 1,000.
Two annotators independently created queries with a third resolving disagreements, reaching a Cohen's $\kappa$ of 0.87 across 47 templates.

We evaluate reasoning segmentation on EndoVis-ReasonSeg and referring segmentation on Ref-EndoVis17 and Ref-EndoVis18~\cite{resurgsam2}.
Performance is measured using region similarity ($\mathcal{J}$) and contour accuracy ($\mathcal{F}$), with FPS evaluated on a single NVIDIA RTX 4090 GPU.
All experiments are repeated over three random seeds and we report mean$\pm$standard deviation.

\noindent\textbf{Reasoning Segmentation Performance.}
Tab.~\ref{tab:endovis_reasonseg_results} compares the performance on EndoVis-ReasonSeg across three reasoning categories.
General-domain methods trained without surgical data perform poorly, with LISA~\cite{lisa} reaching 0.562 $\mathcal{J}\&\mathcal{F}$ on semantic and 0.245 on temporal queries.
This reflects the domain gap, where general multimodal LLMs lack knowledge of instrument functions.
Video-based VISA~\cite{visa} improves over image-based LISA~\cite{lisa} in spatial (0.469 vs.\ 0.207) and temporal (0.329 vs.\ 0.245) reasoning, yet still falls short without surgical priors.
LLM-Seg~\cite{llmseg} achieves the lowest overall performance (0.183 temporal $\mathcal{J}\&\mathcal{F}$) with high variance ($\pm$0.042).
Among surgical methods, ReSurgSAM2~\cite{resurgsam2} reaches 0.593, 0.480, and 0.357 $\mathcal{J}\&\mathcal{F}$.
Compared to ReSurgSAM2~\cite{resurgsam2}, SurgReasonSeg improves $\mathcal{J}\&\mathcal{F}$ by 0.178, 0.237, and 0.297 on semantic, spatial, and temporal reasoning.
All baselines are evaluated from their released weights without fine-tuning on EndoVis-ReasonSeg and without access to the DT, so this comparison measures the end-to-end system rather than the reasoning modules in isolation, which the attribution analysis below addresses.

\noindent\textbf{Efficiency Analysis.}
Most reasoning segmentation methods operate below 2 FPS, with JiT~\cite{jit} at 0.03 FPS from agent-based overhead, while ReSurgSAM2~\cite{resurgsam2} reaches 61.2 FPS but cannot handle implicit queries at all.
Specifically, on a single RTX 4090, DT construction with SurgSAM2, DepthAnything2, and RASO takes 10.6ms, the router 0.8ms, the memory stage 3.2ms, and LoRA inference 34.2ms for $L_{\text{simple}}$ and 48.6ms for $L_{\text{complex}}$.
The memory stage consists of the FAISS lookup, 1.2ms at 500 entries in Tab.~\ref{tab:memory_size} of the supplementary, and the 2.0ms serialization of the $k=3$ retrieved DT segments into the prompt.
The total latency is 48.8ms (20.5 FPS) when $L_{\text{simple}}$ is selected, which accounts for 69\% of queries, and 63.2ms (15.8 FPS) when $L_{\text{complex}}$ is activated.
Averaged over the empirical routing distribution, the expected latency is $0.69\times48.8+0.31\times63.2=53.3$ms, corresponding to 18.8 FPS.
These budgets include DT construction for the incoming frame only, since DT entries for frames already inside the window are cached and reused.
Consequently, the 10.6ms DT construction cost is paid once per frame and does not grow with the window size.
Fig.~\ref{fig:analysis}(a,b) visualizes the trade-off and the per-component latency.

\begin{figure*}[htbp!]
\centering
\includegraphics[width=\linewidth]{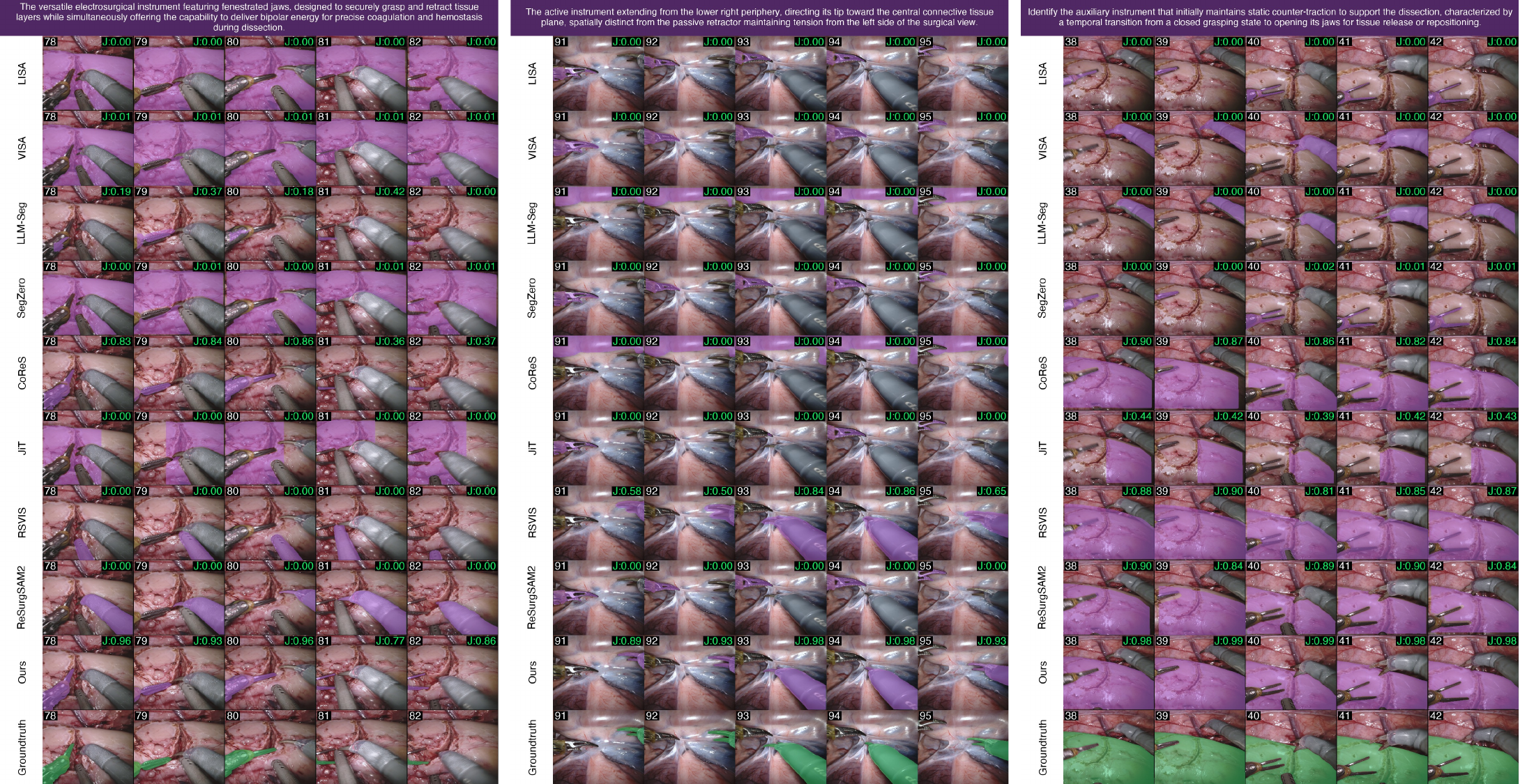}
\caption{
Qualitative comparison of reasoning segmentation on EndoVis-ReasonSeg.
Each row shows the predictions of one method on five consecutive frames, with the frame index at the top left and the $\mathcal{J}$ score at the top right of each frame.
Ground truth is shown in green and predictions in purple, and SurgReasonSeg obtains the highest $\mathcal{J}$ on every frame of the three examples.
}
\label{fig:result}
\end{figure*}

\noindent\textbf{Qualitative Results and Failure Cases.}
In Fig.~\ref{fig:result}, SurgReasonSeg keeps its mask predictions consistent across frames, whereas the other methods produce fragmented masks.
For temporal queries, the adaptive sliding window stabilizes predictions across frames, while methods without temporal modeling produce masks that fluctuate from frame to frame.
For spatial queries, the centroid and depth fields of the DT resolve ambiguities that appearance-only baselines miss in cluttered scenes.
We observe two failure modes of SurgReasonSeg.
The first is a missing candidate, where SurgSAM2 does not instantiate the referent so that no index in the DT can be correct, upper-bounding achievable accuracy at the candidate recall reported in Tab.~\ref{tab:attribution}.
The second is an attribute error, where a RASO label redirects the reasoning chain to a similar instrument or a DepthAnything2 error inverts a near-far relation; the memory occasionally corrects such errors by retrieving a contrasting unsuccessful case.
On the test set, missing candidates account for 29.7\% of the overall failures (consistent with the 8.4\% average candidate miss rate), and attribute errors account for 42.3\%, with the remainder stemming from boundary ambiguity.

\noindent\textbf{Referring Segmentation Performance.}
Tab.~\ref{tab:referring_segmentation_results} compares performance on referring surgical video segmentation, where queries name the target directly.
General-domain methods transfer poorly, with LISA~\cite{lisa} at 0.388, 0.293, and 0.237 $\mathcal{J}\&\mathcal{F}$, while RSVIS~\cite{rsvis} reaches 0.612, 0.684, and 0.657 through video-level consistency and ReSurgSAM2~\cite{resurgsam2} ranks second through its surgical encoder.
SurgReasonSeg consistently achieves the best $\mathcal{J}\&\mathcal{F}$ scores of 0.801, 0.829, and 0.778 across Ref-EndoVis17 (tool), Ref-EndoVis18 (tool), and Ref-EndoVis18 (tissue) datasets.
Compared to ReSurgSAM2~\cite{resurgsam2}, SurgReasonSeg improves performance by 0.024, 0.023, and 0.027 $\mathcal{J}\&\mathcal{F}$ on the three datasets, suggesting that referring segmentation also benefits from accurate DT construction.

The margins on referring datasets (0.024--0.027 $\mathcal{J}\&\mathcal{F}$) are about an order of magnitude smaller than on reasoning queries (0.178--0.297).
When the target is named, ReSurgSAM2 operates close to the segmenter ceiling, so the residual gain reflects DT construction quality rather than reasoning.

\begin{figure}[t!]
\centering
\includegraphics[width=\linewidth]{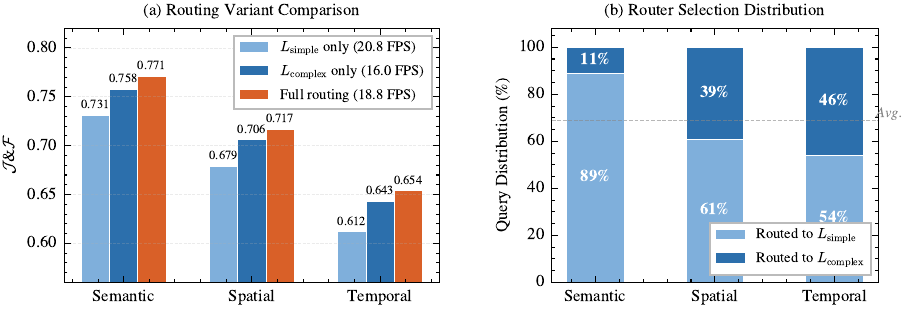}
\caption{
Routing analysis.
The left panel compares $\mathcal{J}\&\mathcal{F}$ across reasoning categories for the single-adapter variants and full routing, with the FPS of each variant in the legend.
The right panel shows the percentage of queries routed to $L_{\text{simple}}$ and $L_{\text{complex}}$ in each category, with the dashed line indicating the overall average.
}
\label{fig:routing}
\end{figure}

\noindent\textbf{Ablation Study.}
Tab.~\ref{tab:ablation_study} reports the ablation of each component.
The baseline without any proposed component, namely the same Qwen3-4B backbone consuming raw frames as visual tokens and decoding a mask without the DT, reaches 0.594, 0.457, and 0.394 $\mathcal{J}\&\mathcal{F}$ at 1.5 FPS, where the low rate reflects pixel-level decoding inside the LLM.
Introducing the DT with ASW and SCM improves $\mathcal{J}\&\mathcal{F}$ to 0.726, 0.663, and 0.590 at 12.5 FPS.
Removing only the adaptive sliding window from the full model gives 0.746, 0.690, and 0.601 at 17.8 FPS, and removing only the surgical context memory gives 0.753, 0.698, and 0.611 at 20.0 FPS.
The adaptive sliding window contributes 0.053 $\mathcal{J}\&\mathcal{F}$ on temporal queries against 0.027 on spatial queries, consistent with spatial queries resting on single-frame relations.
The surgical context memory likewise contributes most on temporal queries, 0.043 $\mathcal{J}\&\mathcal{F}$, by retrieving historical procedural patterns.
Fig.~\ref{fig:analysis}(c) visualizes the contribution of each component across reasoning categories.

We further compare the router with single-adapter variants.
Using only $L_{\text{complex}}$ yields $\mathcal{J}\&\mathcal{F}$ of 0.758, 0.706, and 0.643 at 16.0 FPS, while using only $L_{\text{simple}}$ achieves 0.731, 0.679, and 0.612 at 20.8 FPS, since neither variant runs the router.
The full model with routing achieves 0.771, 0.717, and 0.654 at 18.8 FPS.
The router selects $L_{\text{simple}}$ for 69\% of queries (mostly semantic) and $L_{\text{complex}}$ for 31\% (mostly spatial and temporal).
Its accuracy difference from the always-$L_{\text{complex}}$ variant is 0.013, 0.011, and 0.011 $\mathcal{J}\&\mathcal{F}$, which falls inside the seed-to-seed deviations of $\pm$0.010--0.017 and should therefore be read as parity rather than as an accuracy gain.
The router instead brings a latency benefit at matched accuracy, namely 18.8 against 16.0 FPS, a 17\% throughput increase over always invoking $L_{\text{complex}}$.
On the semantic queries that the router sends to $L_{\text{simple}}$, $L_{\text{complex}}$ scores $0.747\pm0.011$ and $L_{\text{simple}}$ scores $0.750\pm0.010$, so extended reasoning on these queries adds latency without changing accuracy.
The ``no Router'' row of Tab.~\ref{tab:ablation_study} trains a single unspecialized adapter without $R_{\text{length}}$, which is why it reaches only 0.726 on semantic queries and falls below both specialized adapters.
Fig.~\ref{fig:routing} visualizes the routing analysis across reasoning categories.

\begin{table}[t!]
\centering
\caption{\textbf{Matched masks and fine-tuning explain at most 16\% of the margin.}
Masks: the method's own masks or the DT masks of SurgReasonSeg; values are $\mathcal{J}\&\mathcal{F}$ unless noted.
Bold: best; underline: second best; the last block is not ranked.}
\label{tab:attribution}
\footnotesize
\setlength{\tabcolsep}{3pt}
\begin{tabularx}{\columnwidth}{@{}Lcccc@{}}
\toprule
Setting & Masks & Semantic $\uparrow$ & Spatial $\uparrow$ & Temporal $\uparrow$ \\
\midrule
ReSurgSAM2~\cite{resurgsam2} & own & .593 & .480 & .357 \\
ReSurgSAM2~\cite{resurgsam2} & DT & .607 & .492 & .371 \\
LISA~\cite{lisa} & DT & .572 & .224 & .258 \\
ReSurgSAM2, fine-tuned & own & \underline{.621} & \underline{.515} & \underline{.388} \\
Base LLM, zero-shot & DT & .618 & .442 & .379 \\
\midrule
\ours SurgReasonSeg (Ours) & DT & \textbf{.771} & \textbf{.717} & \textbf{.654} \\
\midrule
Oracle index (upper bound) & DT & .852 & .841 & .829 \\
Candidate recall (\%) & DT & 92.4 & 91.8 & 90.5 \\
Held-out templates & DT & .746 & .692 & .628 \\
Cholec80~\cite{twinanda2017endonet} transfer & DT & .721 & .665 & .583 \\
Per-video $\sigma$ (lower is better) & DT & .028 & .035 & .041 \\
\bottomrule
\end{tabularx}
\end{table}

\noindent\textbf{Attribution and Protocol Analysis.}
\label{sec:attribution}
Because segmentation is reframed as index retrieval over SurgSAM2 and RASO outputs, Tab.~\ref{tab:endovis_reasonseg_results} conflates the surgical segmenter with the reasoning modules.
Tab.~\ref{tab:attribution} separates the two by supplying the identical DT masks to ReSurgSAM2 and LISA so that any residual difference is attributable to index selection rather than mask quality.
The fine-tuned row removes the training-data asymmetry, and the oracle row bounds achievable accuracy by always selecting the correct index, providing a direct measure of reasoning quality.
Handing ReSurgSAM2 the identical DT masks lifts it by only 0.014, 0.012, and 0.014 $\mathcal{J}\&\mathcal{F}$, so the masks explain less than 10\% of the margin.
Even fine-tuning it on our training split, the strongest control available, still trails SurgReasonSeg by 0.150, 0.202, and 0.266.
A zero-shot prompted base LLM over the same DT reaches only 0.618, 0.442, and 0.379, showing that neither the masks nor prompting alone suffices.
Measured against the strongest control, masks and surgical training explain 16\%, 15\%, and 10\% of the total margins of 0.178, 0.237, and 0.297, while the reasoning modules explain the remaining 84\%, 85\%, and 90\%.
The oracle bounds of 0.852, 0.841, and 0.829 place SurgReasonSeg at 90\%, 85\%, and 79\% of attainable accuracy, and the candidate recall of 90.5--92.4\% marks the ceiling the retrieval reframing imposes.
On the protocol, the template-disjoint model loses only 0.025--0.026 $\mathcal{J}\&\mathcal{F}$ on the held-out templates and the per-video deviation is 0.028--0.041, both far below the 0.150--0.266 margin over the strongest control, so neither template memorization nor a single sequence carries the effect.
Transfer to Cholec80~\cite{twinanda2017endonet} reaches 0.721, 0.665, and 0.583 without retraining, above every baseline of Tab.~\ref{tab:endovis_reasonseg_results} despite the unseen procedure.

Three further design decisions are ablated in Tab.~\ref{tab:design} of the supplementary, replacing GRPO by supervised fine-tuning, the RL router by a category rule and a supervised classifier, and the VLM judge for $v_j$ by uniform weighting and by a ground-truth IoU upper bound.
The VLM judge agrees with the two human annotators at a Cohen's $\kappa$ of 0.82.

\noindent\textbf{Limitations.}
Accuracy is bounded by the candidate recall of SurgSAM2, since reasoning cannot recover a referent that the segmenter never instantiates.
The window spans only a few frames, so questions that depend on events minutes earlier can be answered only through the memory, and we do not evaluate such procedure-level questions.
On temporal queries whose evidence lies entirely outside the adaptive window, SurgReasonSeg reaches 0.523 $\mathcal{J}\&\mathcal{F}$ through memory retrieval alone.
The memory is validated on EndoVis videos only, so whether it supports continual adaptation across institutions remains to be tested.

\section{Conclusion}

We propose SurgReasonSeg, which brings DT-based reasoning segmentation to surgical video, constructing the DT from surgical foundation models to bridge the gap between general and surgical domains.
Complexity-aware routing between two LoRA adapters adapts computation to query demands, while the adaptive window and surgical context memory provide workflow-aware temporal modeling.
We also contribute the EndoVis-ReasonSeg benchmark.
SurgReasonSeg achieves the best performance across all reasoning and referring benchmarks at 18.8 FPS averaged over queries, and Tab.~\ref{tab:attribution} decomposes the gain into segmenter and reasoning contributions.
Future work targets open surgery and faster LoRA inference, the main per-frame cost.

{
    \small
    \bibliographystyle{ieeenat_fullname}
    \bibliography{main}

@article{attanasio2021autonomy,
  title={Autonomy in surgical robotics},
  author={Attanasio, Aleks and Scaglioni, Bruno and De Momi, Elena and others},
  journal={Annu. Rev. Control Robot. Auton. Syst.},
  volume={4},
  pages={651--679},
  year={2021}
}

@inproceedings{shen2025operating,
  title={Operating room workflow analysis via reasoning segmentation over digital twins},
  author={Shen, Yiqing and Li, Chenjia and Liu, Bohan and others},
  booktitle={MICCAI},
  pages={415--424},
  year={2025}
}

@article{zhou2023text,
  title={Text promptable surgical instrument segmentation with vision-language models},
  author={Zhou, Zijian and Alabi, Oluwatosin and Wei, Meng and others},
  journal={NeurIPS},
  volume={36},
  pages={28611--28623},
  year={2023}
}

@inproceedings{resurgsam2,
  title={{ReSurgSAM2}: Referring segment anything in surgical video via credible long-term tracking},
  author={Liu, Haofeng and Gao, Mingqi and Luo, Xuxiao and others},
  booktitle={MICCAI},
  pages={435--445},
  year={2025}
}

@article{rsvis,
  title={Video-instrument synergistic network for referring video instrument segmentation in robotic surgery},
  author={Wang, Hongqiu and Yang, Guang and Zhang, Shichen and others},
  journal={IEEE Trans. Med. Imaging},
  volume={43},
  pages={4457--4469},
  year={2024}
}

@article{segzero,
  title={{Seg-Zero}: Reasoning-chain guided segmentation via cognitive reinforcement},
  author={Liu, Yuqi and Peng, Bohao and Zhong, Zhisheng and others},
  journal={arXiv:2503.06520},
  year={2025}
}

@inproceedings{visa,
  title={{VISA}: Reasoning video object segmentation via large language models},
  author={Yan, Cilin and Wang, Haochen and Yan, Shilin and others},
  booktitle={ECCV},
  pages={98--115},
  year={2024}
}

@inproceedings{referringsegmentation,
  title={{ReferItGame}: Referring to objects in photographs of natural scenes},
  author={Kazemzadeh, Sahar and Ordonez, Vicente and Matten, Mark and others},
  booktitle={EMNLP},
  pages={787--798},
  year={2014}
}

@article{surgraw,
  title={{SurgRAW}: Multi-agent workflow with chain of thought reasoning for robotic surgical video analysis},
  author={Low, Chang Han and Wang, Ziyue and Zhang, Tianyi and others},
  journal={IEEE Robot. Autom. Lett.},
  volume={11},
  pages={4857--4864},
  year={2026}
}

@inproceedings{lisa,
  title={{LISA}: Reasoning segmentation via large language model},
  author={Lai, Xin and Tian, Zhuotao and Chen, Yukang and others},
  booktitle={CVPR},
  pages={9579--9589},
  year={2024}
}

@inproceedings{jit,
  title={Online reasoning video segmentation with just-in-time digital twins},
  author={Shen, Yiqing and Liu, Bohan and Li, Chenjia and others},
  booktitle={ICCV},
  pages={24698--24706},
  year={2025}
}

@inproceedings{llmseg,
  title={{LLM-Seg}: Bridging image segmentation and large language model reasoning},
  author={Wang, Junchi and Ke, Lei},
  booktitle={CVPR Workshops},
  pages={1765--1774},
  year={2024}
}

@inproceedings{cores,
  title={{CoReS}: Orchestrating the dance of reasoning and segmentation},
  author={Bao, Xiaoyi and Sun, Siyang and Ma, Shuailei and others},
  booktitle={ECCV},
  pages={187--204},
  year={2024}
}

@article{grpo,
  title={{DeepSeekMath}: Pushing the limits of mathematical reasoning in open language models},
  author={Shao, Zhihong and Wang, Peiyi and Zhu, Qihao and others},
  journal={arXiv:2402.03300},
  year={2024}
}

@inproceedings{lora,
  title={{LoRA}: Low-rank adaptation of large language models},
  author={Hu, Edward J and Shen, Yelong and Wallis, Phillip and others},
  booktitle={ICLR},
  year={2022}
}

@inproceedings{dtr1,
  title={Constructing and interpreting digital twin representations for visual reasoning via large language models and reinforcement learning},
  author={Shen, Yiqing and Unberath, Mathias},
  booktitle={ECCV},
  pages={178--194},
  year={2026}
}

@article{position,
  title={Position: Foundation models need digital twin representations},
  author={Shen, Yiqing and Ding, Hao and Seenivasan, Lalithkumar and others},
  journal={arXiv:2505.03798},
  year={2025}
}

@article{surgicalsam2,
  title={Surgical {SAM} 2: Real-time segment anything in surgical video by efficient frame pruning},
  author={Liu, Haofeng and Zhang, Erli and Wu, Junde and others},
  journal={arXiv:2408.07931},
  year={2024}
}

@inproceedings{raso,
  title={Recognize any surgical object: Unleashing the power of weakly-supervised data},
  author={Li, Jiajie and Quaranto, Brian R and Xu, Chenhui and others},
  booktitle={ICLR},
  year={2025}
}

@article{qwen3,
  title={{Qwen3} technical report},
  author={Yang, An and Li, Anfeng and Yang, Baosong and others},
  journal={arXiv:2505.09388},
  year={2025}
}

@inproceedings{fastreasonseg,
  title={Fast reasoning segmentation for images and videos},
  author={Shen, Yiqing and Unberath, Mathias},
  booktitle={CVPR},
  year={2026}
}

@article{ding2025mosformer,
  title={{MoSFormer}: Augmenting temporal context with memory of surgery for surgical phase recognition},
  author={Ding, Hao and Lian, Xu and Unberath, Mathias},
  journal={arXiv:2503.00695},
  year={2025}
}

@article{xue2025omni,
  title={{AdaVideoRAG}: Omni-contextual adaptive retrieval-augmented efficient long video understanding},
  author={Xue, Zhucun and Zhang, Jiangning and Xie, Xurong and others},
  journal={NeurIPS},
  volume={38},
  pages={6094--6117},
  year={2025}
}

@article{depthanything,
  title={Depth Anything {V2}},
  author={Yang, Lihe and Kang, Bingyi and Huang, Zilong and others},
  journal={NeurIPS},
  volume={37},
  pages={21875--21911},
  year={2024}
}

@article{jin2021temporal,
  title={Temporal memory relation network for workflow recognition from surgical video},
  author={Jin, Yueming and Long, Yonghao and Chen, Cheng and others},
  journal={IEEE Trans. Med. Imaging},
  volume={40},
  pages={1911--1923},
  year={2021}
}

@article{trackgpt,
  title={{TrackGPT}--A generative pre-trained transformer for cross-domain entity trajectory forecasting},
  author={Stroh, Nicholas},
  journal={arXiv:2402.00066},
  year={2024}
}

@inproceedings{dtauto,
  title={Towards robust surgical automation via digital twin representations from foundation models},
  author={Ding, Hao and Seenivasan, Lalithkumar and Shu, Hongchao and others},
  booktitle={MICCAI Workshops},
  pages={21--31},
  year={2026}
}

@inproceedings{dtphase,
  title={Towards robust algorithms for surgical phase recognition via digital twin representation},
  author={Ding, Hao and Zhang, Yuqian and Cheng, Wenzheng and others},
  booktitle={MICCAI Workshops},
  pages={119--129},
  year={2025}
}

@article{dtreview,
  title={Characterising the digital twin: A systematic literature review},
  author={Jones, David and Snider, Chris and Nassehi, Aydin and others},
  journal={CIRP J. Manuf. Sci. Technol.},
  volume={29},
  pages={36--52},
  year={2020}
}

@inproceedings{ditto,
  title={{Ditto}: Building digital twins of articulated objects from interaction},
  author={Jiang, Zhenyu and Hsu, Cheng-Chun and Zhu, Yuke},
  booktitle={CVPR},
  pages={5606--5616},
  year={2022}
}

@inproceedings{llamavid,
  title={{LLaMA-VID}: An image is worth 2 tokens in large language models},
  author={Li, Yanwei and Wang, Chengyao and Jia, Jiaya},
  booktitle={ECCV},
  pages={323--340},
  year={2024}
}

@inproceedings{videochatgpt,
  title={{Video-ChatGPT}: Towards detailed video understanding via large vision and language models},
  author={Maaz, Muhammad and Rasheed, Hanoona and Khan, Salman and others},
  booktitle={ACL},
  pages={12585--12602},
  year={2024}
}

@inproceedings{videollama,
  title={{Video-LLaMA}: An instruction-tuned audio-visual language model for video understanding},
  author={Zhang, Hang and Li, Xin and Bing, Lidong},
  booktitle={EMNLP System Demonstrations},
  pages={543--553},
  year={2023}
}

@inproceedings{gsva,
  title={{GSVA}: Generalized segmentation via multimodal large language models},
  author={Xia, Zhuofan and Han, Dongchen and Han, Yizeng and others},
  booktitle={CVPR},
  pages={3858--3869},
  year={2024}
}

@article{twinanda2017endonet,
  title={{EndoNet}: A deep architecture for recognition tasks on laparoscopic videos},
  author={Twinanda, Andru P and Shehata, Sherif and Mutter, Didier and others},
  journal={IEEE Trans. Med. Imaging},
  volume={36},
  pages={86--97},
  year={2017}
}
}

\maketitlesupplementary

\setcounter{table}{0}
\setcounter{figure}{0}
\renewcommand{\thetable}{S\arabic{table}}
\renewcommand{\thefigure}{S\arabic{figure}}
\setcounter{section}{0}
\renewcommand{\thesection}{S\arabic{section}}

This supplementary material reports the adaptive window sensitivity, the memory bank and backbone scaling, design choice ablations, per-video and per-template results, and the Cholec80 transfer subset.
Fig.~\ref{fig:prompts} reproduces the reasoning and judge prompts verbatim, and the judge sees only the overlaid mask and the query, without the ground truth or the reasoning chain.

\begin{figure*}[t!]
\centering
\begin{minipage}[t]{0.48\textwidth}
\hrule\vspace{2pt}
\scriptsize
\begin{verbatim}
[Reasoning Prompt]
You are an expert surgical AI assistant. You will
be given a Digital Twin representation of the
current surgical scene over a short time window,
formatted as a JSON list. Each entry contains an
index, semantic label, average depth, centroid
coordinates, and a bounding box. Identify the
target object that satisfies the user's surgical
query. Output your step-by-step reasoning within
<think> and </think> tags. Finally, output ONLY
the integer index of the target object within
<answer> and </answer> tags.

Query: {query_text}
Digital Twin: {dt_json}
\end{verbatim}
\vspace{1pt}\hrule
\end{minipage}\hfill
\begin{minipage}[t]{0.48\textwidth}
\hrule\vspace{2pt}
\scriptsize
\begin{verbatim}
[VLM Judge Prompt]
You are a highly qualified surgical evaluator.
You are given a video frame from a surgical
procedure, on which a single instance mask is
overlaid in green. You are also given a
descriptive query. Determine whether the
highlighted mask correctly and unambiguously
satisfies the query.
Answer exactly "Yes" if the mask satisfies the
query, or "No" if it does not.

Query: {query_text}
Verdict:
\end{verbatim}
\vspace{1pt}\hrule
\end{minipage}
\caption{The reasoning prompt (left) and the memory quality judge prompt (right) used by SurgReasonSeg.}
\label{fig:prompts}
\end{figure*}

\section{Adaptive Window Sensitivity}

We further evaluate the sensitivity of adaptive sliding window hyperparameters by testing $w_{\text{base}} \in \{1, 2, 3\}$, $w_{\text{query}} \in \{1, 2, 3\}$, and $w_{\text{phase}} \in \{2, 3, 4\}$.
Across these settings, $\mathcal{J}\&\mathcal{F}$ varies by at most 0.015.
Fig.~\ref{fig:dataset} of the main paper shows the resulting window sizes, where semantic queries stay near $w_{\text{base}}$ and temporal queries during phase transitions receive larger windows.

\section{Memory Bank and Backbone Scalability}
\begin{table}[t!]
\centering
\caption{\textbf{$\mathcal{J}\&\mathcal{F}$ of SurgReasonSeg (Ours) peaks at 500 memory entries.}
$\pm$: standard deviation over three seeds; latency is a cost and is not ranked.}
\label{tab:memory_size}
\footnotesize
\setlength{\tabcolsep}{3pt}
\begin{tabularx}{\columnwidth}{@{}cYYYc@{}}
\toprule
& Semantic & Spatial & Temporal & Latency \\
$|\mathcal{M}|$ & $\mathcal{J}\&\mathcal{F}\uparrow$ & $\mathcal{J}\&\mathcal{F}\uparrow$ & $\mathcal{J}\&\mathcal{F}\uparrow$ & (ms) \\
\midrule
0 & \val{.753}{.012} & \val{.698}{.015} & \val{.611}{.019} & 0.0 \\
100 & \val{.762}{.011} & \val{.709}{.014} & \val{.637}{.018} & 0.5 \\
250 & \val{.768}{.011} & \val{.714}{.014} & \val{.649}{.018} & 0.8 \\
\ours 500 & \vb{.771}{.010} & \vb{.717}{.013} & \vb{.654}{.017} & 1.2 \\
1000 & \vu{.770}{.011} & \vu{.716}{.014} & \vu{.653}{.017} & 2.1 \\
\bottomrule
\end{tabularx}
\end{table}

\begin{table}[t!]
\centering
\caption{\textbf{Qwen3-8B adds at most 0.005 $\mathcal{J}\&\mathcal{F}$ over Qwen3-4B at less than half the FPS.}
$\pm$: standard deviation over three seeds; FPS is end-to-end, a cost, and not ranked.}
\label{tab:llm_backbone}
\footnotesize
\setlength{\tabcolsep}{3pt}
\begin{tabularx}{\columnwidth}{@{}lYYYc@{}}
\toprule
LLM & Semantic & Spatial & Temporal & \\
backbone & $\mathcal{J}\&\mathcal{F}\uparrow$ & $\mathcal{J}\&\mathcal{F}\uparrow$ & $\mathcal{J}\&\mathcal{F}\uparrow$ & FPS \\
\midrule
Qwen3-0.6B & \val{.721}{.016} & \val{.668}{.019} & \val{.598}{.024} & 24.7 \\
Qwen3-1.7B & \val{.748}{.013} & \val{.695}{.016} & \val{.631}{.020} & 19.5 \\
\ours Qwen3-4B (Ours) & \vu{.771}{.010} & \vu{.717}{.013} & \vu{.654}{.017} & 18.8 \\
Qwen3-8B & \vb{.774}{.010} & \vb{.720}{.013} & \vb{.659}{.017} & 8.6 \\
\bottomrule
\end{tabularx}
\end{table}

\noindent\textbf{Memory Bank Scalability.}
Tab.~\ref{tab:memory_size} evaluates the effect of surgical context memory bank size on performance.
Without memory ($|\mathcal{M}|=0$), temporal reasoning drops by 0.043 $\mathcal{J}\&\mathcal{F}$ compared to the full model, showing that historical procedural context matters most for temporal queries.
The impact on semantic reasoning is smaller ($-$0.018 $\mathcal{J}\&\mathcal{F}$), as semantic queries are often resolved from single-frame DT attributes.

Performance improves consistently up to 500 entries and slightly decreases at 1,000.
Beyond 500 entries, retrieval noise from marginally similar cases outweighs the added coverage, suggesting that the quality-aware scoring with $v_j$ becomes less discriminative.
At 500 entries, the FAISS lookup takes 1.2ms, under 2.5\% of the total inference time.
At 1,000 entries, the retrieval latency of 2.1ms remains compatible with real-time operation.
All memory results come from EndoVis videos, so multi-institution deployment remains to be tested.

\noindent\textbf{LLM Backbone.}
Tab.~\ref{tab:llm_backbone} compares different Qwen3 model sizes as the base LLM backbone.
Scaling from 0.6B to 4B parameters yields consistent improvements across all reasoning categories, with the largest gains in temporal reasoning ($+$0.056 $\mathcal{J}\&\mathcal{F}$), as larger models better capture complex procedural dependencies.

However, scaling to 8B provides only marginal improvements ($+$0.003--0.005 $\mathcal{J}\&\mathcal{F}$) while lowering the FPS from 18.8 to 8.6, below the real-time threshold.
This saturation suggests that, once visual perception is offloaded to the foundation models that build the DT, a 4B model is large enough for the reasoning step.
We therefore use Qwen3-4B, which balances reasoning accuracy against real-time inference.

\section{Design Choice Ablations}
\begin{table}[t!]
\centering
\caption{\textbf{Every design choice of SurgReasonSeg (Ours) outperforms its replacement, and the VLM judge is within 0.004 of the ground-truth IoU bound.}
Adapters: RL or supervised fine-tuning (SFT); Router: RL, query-category rule, or supervised classifier; $v_j$: VLM judge, uniform weighting, or ground-truth IoU; gray: as in the full model.
Values are $\mathcal{J}\&\mathcal{F}$; bold: best; underline: second best; the ground-truth IoU row is an upper bound and is not ranked.}
\label{tab:design}
\footnotesize
\setlength{\tabcolsep}{3pt}
\begin{tabularx}{\columnwidth}{@{}lllYYY@{}}
\toprule
Adapters & Router & $v_j$ & Semantic $\uparrow$ & Spatial $\uparrow$ & Temporal $\uparrow$ \\
\midrule
SFT & \gs{RL} & \gs{Judge} & .742 & .688 & .615 \\
\gs{RL} & Rule & \gs{Judge} & .731 & .706 & .643 \\
\gs{RL} & Classifier & \gs{Judge} & \underline{.762} & \underline{.710} & \underline{.648} \\
\gs{RL} & \gs{RL} & Uniform & .758 & .702 & .640 \\
\midrule
\ours RL & RL & Judge & \textbf{.771} & \textbf{.717} & \textbf{.654} \\
\midrule
\gs{RL} & \gs{RL} & GT IoU & .775 & .721 & .658 \\
\bottomrule
\end{tabularx}
\end{table}

Tab.~\ref{tab:design} ablates three design decisions independently.
Replacing GRPO with supervised fine-tuning on the same rollouts tests whether RL is required or whether the reward merely selects an output length that supervision could imitate.
Replacing the RL router with a category rule and a supervised classifier tests whether the router must be learned against a latency-aware reward.
Varying $v_j$ between uniform weighting, a ground-truth IoU bound, and the external judge makes the cost of an imperfect judge measurable.
The VLM judge agrees with the human annotators at a Cohen's $\kappa$ of 0.82, which explains why the performance with the VLM judge approaches that of the ground-truth IoU upper bound.

\section{Per-Video and Per-Template Breakdown}
Tab.~\ref{tab:attribution} of the main paper reports the aggregate per-video deviation, and Tab.~\ref{tab:pervideo} resolves it into the individual test sequences.
No single video carries the reported margins.
The weakest sequence still reaches 0.728, 0.663, and 0.591 $\mathcal{J}\&\mathcal{F}$, exceeding the strongest control of Tab.~\ref{tab:attribution}, ReSurgSAM2 fine-tuned on our training split, by 0.107, 0.148, and 0.203, so the ordering holds sequence by sequence rather than only in aggregate.
The 0.075--0.110 spread tracks the source dataset, with the three EndoVis17 sequences below the four EndoVis18 ones, consistent with the older subset's lower annotation density.
Tab.~\ref{tab:templates} reports the template inventory and the ten templates held out for the template-disjoint split.
The ten held-out templates contribute 205 of the 961 test samples, on which the template-disjoint model scores 0.746, 0.692, and 0.628, only 0.025 to 0.026 below the full model on the standard split.
Unseen phrasings therefore yield comparable accuracy to seen ones, indicating that the model resolves queries through the DT rather than through template memorization.

\begin{table}[t!]
\centering
\caption{\textbf{Every test video stays above the strongest control of the main paper.}
Per-sequence $\mathcal{J}\&\mathcal{F}$ of SurgReasonSeg.}
\label{tab:pervideo}
\footnotesize
\setlength{\tabcolsep}{3pt}
\begin{tabularx}{\columnwidth}{@{}lcYYY@{}}
\toprule
Sequence & Source & Semantic & Spatial & Temporal \\
\midrule
seq-1 & EndoVis17 & .728 & .663 & .591 \\
seq-2 & EndoVis17 & .743 & .682 & .613 \\
seq-3 & EndoVis17 & .763 & .707 & .642 \\
seq-4 & EndoVis18 & .776 & .723 & .661 \\
seq-5 & EndoVis18 & .789 & .740 & .680 \\
seq-6 & EndoVis18 & .795 & .747 & .689 \\
seq-7 & EndoVis18 & .803 & .757 & .701 \\
\midrule
Mean & All & .771 & .717 & .654 \\
Std.\ dev. & All & .028 & .035 & .041 \\
\bottomrule
\end{tabularx}
\end{table}

\begin{table}[t!]
\centering
\caption{\textbf{Ten of the 47 templates are held out and cover 205 test samples.}
The last row is the template-disjoint model; All is weighted by sample count.}
\label{tab:templates}
\footnotesize
\setlength{\tabcolsep}{3pt}
\begin{tabularx}{\columnwidth}{@{}Lcccc@{}}
\toprule
& Semantic & Spatial & Temporal & All \\
\midrule
Templates & 20 & 15 & 12 & 47 \\
Held-out templates & 4 & 3 & 3 & 10 \\
Test samples, held out & 85 & 68 & 52 & 205 \\
$\mathcal{J}\&\mathcal{F}$, held out & .746 & .692 & .628 & .698 \\
\bottomrule
\end{tabularx}
\end{table}

\section{Cholec80 Transfer Subset}
The transfer row of Tab.~\ref{tab:attribution} uses a reasoning segmentation subset built from Cholec80, a laparoscopic cholecystectomy corpus disjoint from EndoVis in procedure and instruments.
We sample 10 sequences and segment them into clips of 20 to 40 frames at instrument-change boundaries, following the same protocol as EndoVis-ReasonSeg, and author 420 implicit queries from the held-out template pool (Cohen's $\kappa=0.85$).
The model is evaluated without retraining and with the memory bank populated only from EndoVis training videos, so the setting measures cross-procedure transfer rather than adaptation.
Accuracy drops by 0.050, 0.052, and 0.071 $\mathcal{J}\&\mathcal{F}$ relative to the in-domain result, consistent with the instrument vocabulary shift, yet the transferred model still exceeds every baseline of Tab.~\ref{tab:endovis_reasonseg_results}.

\end{document}